\documentclass[10pt,letterpaper]{article}
\usepackage[top=0.85in,left=2.75in,footskip=0.75in]{geometry}

\usepackage{fullpage}

\usepackage[utf8]{inputenc}

\usepackage{textcomp,marvosym}

\usepackage{amsmath,amssymb}

\usepackage{cite}

\usepackage{nameref,hyperref}

\usepackage[right]{lineno}

\usepackage{microtype}
\DisableLigatures[f]{encoding = *, family = * }

\usepackage{rotating}

\usepackage[aboveskip=1pt,labelfont=bf,labelsep=period,justification=raggedright,singlelinecheck=off]{caption}

\makeatletter
\renewcommand{\@biblabel}[1]{\quad#1.}
\makeatother

\date{}
\usepackage{multirow}
\usepackage{lastpage,fancyhdr,graphicx}
\usepackage{epstopdf}

\usepackage{algorithm}
\usepackage{algpseudocode}
\usepackage{verbatim}

\newcommand{\OO}{\mathcal{O}}
\newcommand{\CC}{\bf{C}}
\newcommand{\JJ}{\mathcal{J}}

\newcommand{\mb}[1]{\mbox{\boldmath $#1$}}

\newcommand{\bx}{\mb{x}}

\begin{document}

%
%

\begin{flushleft}
{\Large
\textbf\newline{Multilevel Weighted Support Vector Machine  for Classification on Healthcare Data with Missing Values}
}
\newline
\\
Talayeh Razzaghi\textsuperscript{1},
Oleg Roderick\textsuperscript{2},
Ilya Safro\textsuperscript{1*},
Nicholas Marko\textsuperscript{2},
\\
\bigskip
\bf{1} School of Computing, Clemson University, Clemson, SC, USA
\\
\bf{2} Department of Data Science, Geisinger Health System, Danville, PA, USA
\\
\bigskip

* isafro@clemson.edu

\end{flushleft}


\section*{Abstract}
\begin{abstract}
		This work is motivated by the needs of predictive analytics on healthcare data as represented by Electronic Medical Records. Such data is invariably problematic: noisy, with missing entries, with imbalance in classes of interests, leading to serious bias in predictive modeling. Since standard data mining methods often produce poor performance measures, we argue for development of specialized techniques of data-preprocessing and classification. In this paper, we propose a new method to simultaneously classify large datasets and reduce the effects of missing values. It is based on a multilevel framework of the cost-sensitive SVM and
	the expected maximization imputation method for missing values, which relies on iterated regression
	analyses. We compare classification results of multilevel SVM-based algorithms on public benchmark datasets with imbalanced classes and missing values as well as real data in health applications, and show that our multilevel SVM-based method produces fast, and more accurate and robust classification results.
\end{abstract}


\section{Introduction}  \label{sec:intro}

Modern healthcare can be characterized as \textit{personalized}, \textit{evidence-driven} and \textit{model-assisted} \cite{foldy2014national}. As healthcare industry is becoming more integrated with data science, planners and practitioners have to continuously choose the best available machine learning methods to use on medical data that is inherently sparse, noisy, and scanned for rare events more often that for the norm.

In the clinical environment, decisions made based on predicting risks and positive outcomes should be ideally supported by statistical learning models. These models can be seen as either a simplified risk-assessment model \cite{haas2013risk}, or a sophisticated machine learning method \cite{plis2014machine,woolery1994machine}. In either case, it is based on a query of relevant clinical and operational history. While prediction of real-valued health metric over time is of interest to the healthcare domain, it more properly belongs to the field of medical simulation models.


Due to the nature of big data in healthcare, the role of data analysis, particularly classification methods, is critical to support better decision for personalized medicine, that is, decision-making with awareness that patients can be classified into groups based on their personal characteristics and the patterns observed in patient-provider-insurer interactions, and that patients from different groups will have different responses to treatment and different risk outcomes. Thus, if we describe the most common task of predictive analytics for healthcare in one informal sentence, it would be: solving classification problems on clinical data using specialized pre-processing and specialized predictive algorithms.

Comprehensive medical information comes in multiple categories (that can be stored in multiple databases, with different formats and rules of access). Categories include: biometric information, medical codes referring to clinical transactions, insurance claims and payments, results of laboratory tests, narratives such as doctors' notes, socioeconomic data characterizing life conditions and choices of individual patients, and molecular data (genomics, proteomics, metabolomics). Our work is focused on the categories items placed earlier in this list. They are considered 'traditional medical information', are recorded for millions, rather than thousands of patients, and are best source material for study of large patient populations. Due to considerations of patient privacy, and the proprietary nature of electronic medical records \cite{larson2004survey}, the databases cannot be queried continuously. Every instance of data acquisition and integration is a separate effort that is cost-effective only when the resulting predictive model shows high quality. Thus, progress in evidence-driven healthcare depends on how well state-of-the art algorithms of machine learning are adapted to clinical data.

We note that classical mathematical, and computer science issues, such as scalability, or convergence rate are rarely a major issue for healthcare applications. Instead, an algorithm is ranked based on its ability to process raw medical data, with such problematic features as sparsity, missing entries, noise and imbalanced outputs. Because of the encounter nature of patient-provider interaction, medical data is inherently sparse: when a clinical encounter occurs, the number of and contents of labels attached to it vary widely \cite{snomed2011systematized}; outside of an encounter, the state of the patient is unknown. The outcomes of interest in classification problems are imbalanced, because, as a rule, healthcare analytics is motivated by rare events such as healthcare emergencies, severe chronic conditions, gaps and bottlenecks in access to care.

This work was prompted by several projects completed with the Division of Applied Research and Clinical Informatics, Dept. of Data Science; Geisinger Health System. For the first motivating example (Example 1, see Section \ref{sec:numres}), we use our 2014 feasibility study \cite{RoderickMEDAI} of merging insurance information (6 aggregate features, based on the history of claims and payments) together with clinical encounter information (10-20 features chosen by hand from patient biometrics, medications and diagnostic codes). The goal of the initial study was to predict the financial risk for a particular patient (a common metric in insurance practice, derived as a ratio of individual expenses and average expenses for a large demographic group). 
We attempted to use a standard clustering technique, k-nearest neighbors with empirically selected weighting, to achieve the basic results before we developed the proposed method.
Unfortunately, the results were very unsatisfactory: the chance of mis-categorization (in one-against all binary classification) was close to 50\% for all risk groups. Intuitive explanations such as “some patients have entered a high-risk state that is not yet reflected in their financial information” could not be formally verified or used to explain the poor performance, which prompted interest in using the more advanced machine learning methods. 

For Example 2 (see Section \ref{sec:numres}), we use our preliminary investigation of patients' response to public outreach \cite{RoderickFLU}, such as annual flu awareness campaigns. We included basic demographic and clinical information on patients targeted by 35 identically organized campaigns.
(the features included in the data were: age, sex and BMI of the patient, and binary variables identifying whether the patient was assigned the most commonly occurring medication codes and prescribed the most commonly occuring medications).
The data was used to build a model predicting whether a given patient is likely to respond to the reminder, or to choose not to get vaccinated, or use a different provider. Again, our initial core predictive model was standard: logistic regression with empirically selected weighting of training data which is widely used in healthcare informatics.

We intend to show that in each case, the predictive models are made more effective with the use of an advanced machine learning algorithm developed with awareness of sparsity and class skewness (imbalance) in data.


\section{Methods}
Our study was approved by Geisinger’s Institutional Review Board. Information from individual electronic medical records was de-identified prior to use in the study.

Support vector machines (SVM) are among the most well-known optimization-based supervised learning methods, originally developed for binary classification problems \cite{vapnik2000nature}. The main idea of SVM is to identify a decision boundary with maximum possible margin between the data points of each class.
Training nonlinear SVMs is often a time consuming task when the data is large. This problem becomes extremely sensitive when the model selection techniques are applied.
Requirements of computational resources, and storage are growing rapidly with the number of data points, and the dimensionality, making many practical classification problems less tractable. In practice, when solving SVM, there are several parameters that have to be tuned. Advanced methods, such as the grid search and the uniform design for tuning the parameters, are usually implemented using iterative techniques, and the total complexity of the SVM strongly depends on these methods, and on the quality of the employed optimization solvers such as  \cite{chang2011libsvm}.

In this paper, we focus on SVMs that are formulated as the convex quadratic programming (QP). Usually, the complexity required to solve such SVMs is between $\OO(n^2)$ to $\OO(n^3)$ \cite{graf2004parallel}. For example, the popular QP solver implemented in LibSVM \cite{chang2011libsvm} scales between $\OO(n_{f}{n_{s}}^2)$ to $\OO(n_{f}{n_{s}}^3)$ subject to how efficiently the LibSVM cache is employed in practice, where $n_f$, and $n_s$ are the numbers of features, and samples, respectively.

Typically, the gradient descent methods achieve good performance results on such models, but still tend to be
very slow for large-scale data (when effective parallelism is hard to achieve). Several works have recently addressed this problem. Parallelization usually splits the large set into smaller subsets and then performs a training to assign data points into different subsets \cite{collobert2002parallel}. In \cite{graf2004parallel}, a parallel version of the Sequential Minimal Optimization (SMO) was developed to accelerate the solution of QP. Although parallelizations over the full data sets often gain good performance, they can be problematic to implement due to the dependencies between variables, which increases communication. Moreover, although specific types of SVMs might be appropriate for parallelization (such as the Proximal SVM \cite{tveit2003multicategory}), the question of their practical applicability for high-dimensional datasets still requires further investigation.
Another approach to accelerate the QP is chunking \cite{Joachims1999,catak2013cloudsvm}, in which the optimization problem is solved iteratively on the subsets of training data until the global optimum is achieved.
The SMO is among the most popular methods of this type \cite{platt1999fast}, which scales down the chunk size to two vectors. Shrinking to identify the non-support vectors early, during the optimization, is another common method that significantly reduces the computational time \cite{Joachims1999,chang2011libsvm,collobert2002torch}. Such techniques can save substantial amounts of storage when combined with caching of the kernel data.
Digesting is another successful strategy that ``optimizes subsets of training data closer to completion before adding new data`` \cite{decoste2002training}.
\emph{Summarizing computational and EMR problems mentioned above, we note that being highly flexible and parametrizable to be applied on a variety of complex manifolds, applications of SVMs on large-scale healthcare data without significant decrease in time complexity can be extremely expensive.}

Let us formally describe a supervised classification problem on data consisting of real-valued variables and categorical variables converted into binaries.
Given a training set $\mathcal{J}=\{(x_i, y_i)\}_{i=1}^l$, that is a set of data points with known labels, where $(x_i, y_i)~\in~\mathbb{R}^{n+1}$, and  $l$ and $n$ are the numbers of data points and features, respectively, and $y_i \in \{-1,1\}$ denotes the class label for each data point $i$ in $\mathcal{J}$. We denote by $\CC^-$ and $\CC^+$, the "majority'' (points with $y_i=-1$) and ``minority`` (points with $y_i=+1$) classes respectively such that $\mathcal{J}=\CC^+ \cup \CC^-$.

\subsection{Support Vector Machines}
The optimal SVM classifier is determined by the parameters $w$ and $b$ through solving the convex optimization problem 
\begin{subequations}\label{softmarginSVM}
\begin{align}
              \min &\ \ {\frac{1}{2}\ \| w \|^2+C \sum_{i=1}^{l}\xi_i} \\
               \text{s.t.}&\ \               y_i(w^T\phi(x_i)+b)\geq1-\xi_i &  \;i = 1, \ldots, l\\
            & \ \  \xi_i\geq0  &     \;i = 1, \ldots, l
\end{align}
\end{subequations}
\noindent where $\phi$ maps training instances $x_i$ into a higher dimensional space, $\phi: \mathbb{R}^{n} \to \mathbb{R}^{m}$  ($m\geq n$). The term slack variables $\xi_i$ ($i \in \{1, \ldots, l\}$) in the objective function is used to penalize misclassified points. This approach is also known as {\em soft margin} SVM. The magnitude of penalization is controlled by the parameter $C$. Many existing algorithms (such as SMO, and its implementation in LIBSVM tool  \cite{chang2011libsvm} that we use) solve the Lagrangian dual problem instead of the primal formulation, which is a popular strategy due to its faster and more reliable convergence.

\subsection{Weighted Support Vector Machines}

Imbalanced classification tasks (when the sizes of classes are very different) are another major problem that, in practice, can lead to poor performance measures \cite{tang2009svms}.  Imbalanced learning is a significant emerging problem in many areas, including medical diagnosis  \cite{lo2008learning,mazurowski2008training,li2010learning}, face recognition \cite{kwak2008feature}, bioinformatics \cite{batuwita2009micropred}, risk management \cite{ezawa1996learning,groth2011intraday}, and manufacturing \cite{su2007evaluation}. Many standard SVM algorithms often tend to misclassify the data points of the minority class. One of the most well-known techniques to deal with imbalanced data is the cost-sensitive learning (CSL). The CSL addresses imbalanced classification problems through different cost matrices. The adaptation of cost-sensitive learning with the regular SVM is known as \emph{weighted support vector machine} (WSVM \cite{veropoulos1999controlling}, also termed as Fuzzy SVM) \cite{lin2002fuzzy}. The main idea is to consider weighting scheme in learning such that the WSVM algorithm builds the decision hyperplane based on the relative contribution of data points in training. In contrast to the standard SVM, the penalization costs are different for the positive ($\CC^+$) and negative ($\CC^-$) classes:
\begin{subequations}\label{eq:wsvm}
\begin{align}
              \min &\ \ {\frac{1}{2}\ \| w \|^2+C^+ \sum_{\{i|y_i=+1\}}^{n_+}\xi_i+C^-\sum_{\{j|y_j=-1\}}^{n_-}\xi_j}\\
              \textrm{s.t.}&\ \  y_i(w^T\phi(x_i)+b)\geq 1-\xi_i \hspace{20pt}  i = 1, \ldots, l\\
&\ \ \xi_i\geq0 \hspace{100pt}  i = 1, \ldots, l,
\end{align}
\end{subequations}
where $C^+$, and $C^-$ are the parameters associated with the positive, and negative classes, which assign different importance weights to each data class.
The formulations (\ref{softmarginSVM}) and (\ref{eq:wsvm}) are solved through the Karush-Kuhn-Tucker conditions.

The Gaussian kernel function (radial basis function, RBF) defined as
\begin{equation}
       k(\bx_i,\bx_j)=\exp(-\gamma \|\bx_i-\bx_j\|^2),  \gamma \geq 0,
 \end{equation}
is used in the dual formulation of (W)SVMs. This kernel has been confirmed as the most successful for the UCI benchmark in multiple studies.
Parameter tuning is required to set optimal or near optimal $C$, $C^+$, $C^-$, and kernel function parameters (e.g. bandwidth parameter for RBF kernel function) to achieve good results for (W)SVM. This process becomes problematic and time-consuming particularly when the size of data is very large. Hence we aim to develop an efficient and effective classification method, called the Multilevel (W)SVM, that is scalable and works with imbalanced healthcare data.


\subsection{Multilevel Support Vector Machines}
The proposed algorithm belongs to the family of multilevel optimization strategies \cite{brandt:optstrat} whose goal is to approximate the system at multiple scales of coarseness and to obtain a final solution by combining the information from different scales. The multilevel framework for SVM \cite{razzaghi2015scalable} scales efficiently for large classification problems whose hierarchy of coarser representations is constructed based on the approximated $k$-nearest neighbors graphs (A$k$NN).
We note that the exact nearest neighbor graph methods are rather computationally expensive due to construction of the k-NN graph structure. There is a lack of  exact and scalable nearest neighbor search algorithms with good performance, when data is high-dimensional. Several attempts have been made to propose an approximate search \cite{muja2014scalable,muja_flann_2009}, in which not all the neighbors obtained are exact, but still generally close to the exact neighbors.
The multilevel support vector machine method consists of three main phases, namely, coarsening, coarse support vector initial learning, and uncoarsening.

\paragraph{The coarsening phase.}

The coarsening algorithms are the same for both $\CC^+$, and $\CC^-$, so we provide only one of them. Given a class of data points $\CC$, the coarsening begins with a construction of an approximated $k$-nearest neighbors (A$k$NN) graph $G=(V,E)$, where $V=\CC$, and $E$ are the edges of A$k$NN.
A gradual coarsening of the training set is constructed using fast point selection method \cite{saaddim} in A$k$NN graph. (In fact, this version of coarsening is a simplified coarsening developed for combinatorial optimization problems on graphs such as in \cite{RonSB11,SafroRB06}.)

The goal is to select a set of representative points $\hat{V}$ for the next-coarser problem, where $\vert\hat{V}\vert \geq Q \vert V\vert$, and $Q$ is the parameter for the size of the coarse level graph. In practice, it can often be done by selecting a maximal independent set of points such as in \cite{saaddim}. However, we found that ensuring a slightly denser uniform coverage of the points can lead to much better results than finding an independent set of points (nodes in A$k$NN) as was suggested in \cite{saaddim}.
(An independent set is a set of vertices in a graph when no two vertices are connected by an edge.)
Thus, we extended the set of coarse points by setting a parameter for the minimum number of points that in our experiments was set to 50\% of the fine data points. The second requirement for $\hat{V}$ is that it has to be a dominating set of $V$. (The dominating set of nodes is a subset of $V$ such that each vertex in $V$ is either in this set or adjacent to one or more vertices in it.)

The coarsening for class $\CC$ is presented in Algorithm \ref{alg1}. The algorithm consists of several iterations of independent set of $V$ selections that are complementary to already chosen sets. We begin with choosing a random independent set (line 2) using the greedy algorithm. It is eliminated from the graph, and the next independent set is chosen and added to $\hat{V}$ (lines 5-11).
For imbalanced cases, when WSVM is used, we avoid of creating very small coarse problems for $\CC^-$. Instead, already very small class is continuously replicated across the rest of the hierarchy if $\CC^+$ still requires coarsening. We note that this method of coarsening will reduce the degree of skewness in the data and make the data approximately balanced at the coarsest level.
The multilevel framework recursively calls the coarsening process until it creates a hierarchy of $r$ coarse representations $\{\JJ_i\}_{i=0}^r$ of $\JJ$. At each level of this hierarchy, the corresponding A$k$NNs' $\{G_i=(V_i,E_i)\}_{i=0}^r$ are saved for future use at the uncoarsening phase. The corresponding data and labels at level $i$ is denoted by $(X_i,Y_i) \in \mathbb{R}^{k\times{(n+1)}}$, where $|X_i|=k$.

\begin{table}[t]
\begin{tabular}{ll}
\begin{minipage}{0.5\textwidth}
\begin{algorithm}[H]
\caption{The Coarsening}
\label{alg1}
\begin{algorithmic}[1]
\State {\bf Input:} $G=(V,E)$ for class $\CC$
 \State  $\hat{V} \gets $ select maximal independent set in $G$
 \State   $\hat{U} \gets V \setminus \hat{V}$
\While{$\vert \hat{V}\vert<Q\cdot |V|$}

\While{$\hat{U} \neq \emptyset$}

     \State  randomly pick $i \in \hat{U}$
     \State $\hat{U} \gets \hat{U} \setminus \{i\}$
     \State $\hat{U} \gets \hat{U} \setminus \{\text{neighbors of } i \text{ in } \hat{U}\}$
     \State $\hat{V} \gets \hat{V} \cup \{i\}$

\EndWhile

  \State   $\hat{U} \gets V \setminus \hat{V}$

\EndWhile\\
\Return $\hat{V}$
\end{algorithmic}
\end{algorithm}
\end{minipage}
&
\begin{minipage}{0.5\textwidth}
    \includegraphics[width=0.98\textwidth]{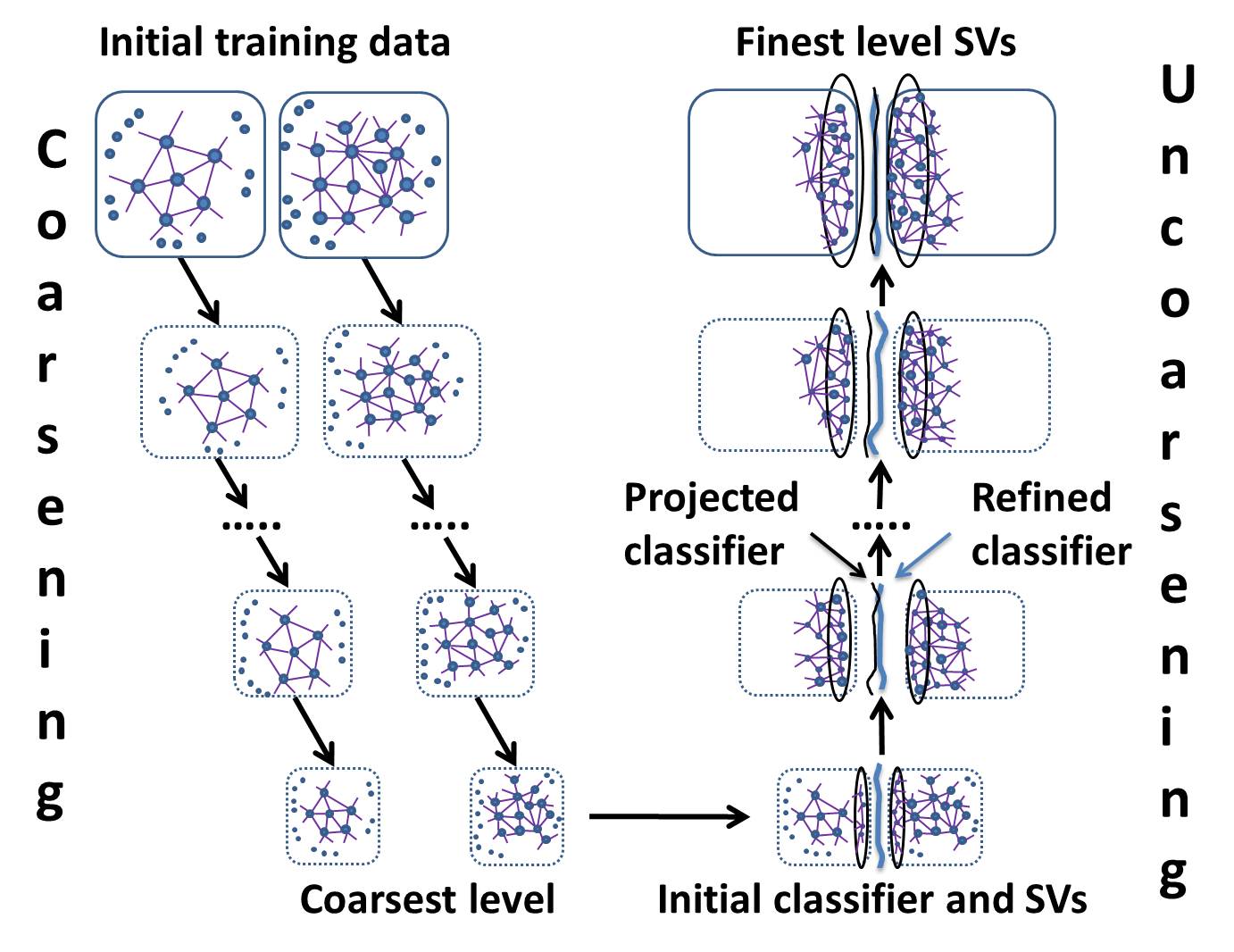}
    \captionof{figure}{The multilevel SVM framework consists of three phases: gradual training set
coarsening, coarsest support vectors’ learning, and gradual support vectors’ refinement (uncoarsening). Pairs of A$k$NN graphs correspond to two classes of learning.}\label{fig:ml}
\end{minipage}
\end{tabular}
\end{table}

 \paragraph{Supervised support vector initial learning.} After the hierarchy is created, the support vectors learning is performed at the coarsest level, where the number of data points is sufficiently small. \\
At the coarsest level $r$, when $|\JJ_r| << \JJ$, we can apply an exact algorithm for training the coarsest classifier. Typically, the size of the coarsest level depends on the computational resources. However, for the (W)SVM problems, one can also consider some criteria of the separability between $\CC_{\emph r}^+$, and $\CC_{\emph r}^-$ \cite{wang2008feature}, i.e., if a fast separability test exists or additional data properties are available. We used the simplest criterion bounding $\JJ_r$ to 500. Processing the coarsest level includes an application of the uniform design (UD) \cite{huang2007model} model selection to improve the quality of classifiers.
The nested UD search is an efficient method used for automatic model selection for SVMs. This method is applied to select the candidate set of parameter combinations and carry out a k-fold cross-validation to evaluate the quality of each parameter combination.

      \paragraph{The uncoarsening phase.} Support vectors, and classifier are projected throughout the hierarchy from the coarsest to the finest levels. At each level, a solution to the current fine level is updated and optimized based on the solution of the previous coarse level. The locally optimal support vectors are obtained by gradual refinement of the support vectors projected from the coarse level.

Given the solution of coarse level $i+1$ (the set of support vectors $S_{i+1}$, and parameters $C_{i+1}$, and $\gamma_{i+1}$), the primary goal of the refinement is to update and optimize this solution for the current fine level $i$. Unlike many other multilevel algorithms, in which the inherited coarse solution contains projected variables only, in our case, we initially inherit not only the coarse support vectors (the solution that can represent the whole training set \cite{syed1999incremental,fung2002incremental}) but also parameters for model selection. This is because the model selection is an extremely time-consuming component of (W)SVM, and can be prohibitive at fine levels. However, at the coarse levels, when the problem is much smaller than the original, we can apply much heavier methods for model selection with almost no loss in the running time of the framework.
In particular, at each level of ML(W)SVM, after updating the training set and before running SVM, the UD is performed on the training data. Because the data might be imbalanced, we select the optimal parameter set with respect to the maximum G-mean value. The optimal $C_{i+1}$, $\gamma_{i+1}$ of previous level are used as the initial $C_i$ and $\gamma_i$ at level $i$, and they will be updated at each level based on the new training set. The $C_i$ and $\gamma_i$ that result in higher G-mean will be selected as optimal or near-optimal parameters at all levels.

The refinement is presented in Algorithm \ref{alg3}. The coarsest level is solved exactly and reinforced by the model selection (lines 2-5). If $i$ is one of the intermediate levels, we build the set of training data $data^{(i)}_{train}$ by inheriting the coarse support vectors $S_{i+1}$ and adding to them some of their approximated nearest neighbors at level $i$ (lines 6-7) (in our experiments, usually not more than 5). If the size of $data^{(i)}_{train}$ is still small enough (relatively to the existing computational resources, and the initial size of the data) for applying model selection, and solving SVM on the whole $data^{(i)}_{train}$, then we use coarse parameters $C_{i+1}$, and $\gamma_{i+1}$ as initializers for the current level, and retrain (lines 9-10,19). Otherwise, the coarse $C_{i+1}$, and $\gamma_{i+1}$ are inherited in $C_{i}$, and $\gamma_{i}$ (line 12). Then, being large for direct application of the SVM, $data^{(i)}_{train}$ is clustered into $K$ clusters, and pairs of $P$ nearest opposite clusters are retrained, and contribute their solutions to $S_i$ (lines 15-17). The number of $K$ is determined in as
                           \begin{equation}
                                  K = |{data}^{(i)}_{train}| / Q_{dt}.
                           \end{equation}
\emph{We note that cluster-based retraining can be done in parallel, as different pairs of clusters are independent. Moreover, the total complexity of the algorithm does not suffer from reinforcing the cluster-based retraining with model selection.}

\begin{algorithm}
\caption{The Refinement at level $i$}
\label{alg3}
\begin{algorithmic}[1]
\State {\bf Input:} $\JJ_i, S_{i+1}, C_{i+1}, \gamma_{i+1}$
\If {$i$ is the coarsest level}
 \State Calculate the best ($C_{i}$, $ \gamma_{i}$) using UD
  \State $S_i \leftarrow$ Apply SVM on $X_i$
  \EndIf

  \State Calculate nearest neighbors $N_i$ for support vectors $S_{i+1}$ from the existing A$k$NN $G_i$
                           \State  ${data}^{(i)}_{train} \gets S_{(i+1)} \cup N_i$
                                 \If {$|{data}^{(i)}_{train}| < Q_{dt}$ }
                                       \State $C^O  \gets C_{i+1}$; $\gamma^{O} \gets \gamma_{i+1}$
                                        \State Run UD using the initial center $(C^O, \gamma^{O})$

                                 \Else
                                       \State $C_i \leftarrow C_{i+1}$; $\gamma_i \leftarrow \gamma_{i+1}$
                                  \EndIf
                                  \If {$|{data}^{(i)}_{train}| \geq Q_{dt}$ }\par
                                 \State Cluster ${data}^{(i)}_{train}$ into $K$ clusters
                                 \State $\forall k\in K$ find $P$ nearest opposite-class clusters
               \State $S_i \leftarrow$ Apply SVM on pairs of nearest clusters only

                                  \Else
                                       \State $S_i \leftarrow$ Apply SVM directly on ${data}^{(i)}_{train}$
                                  \EndIf
                                  \State {\bf Return} $S_i$, $C_i$, $\gamma_i$
\end{algorithmic}
\end{algorithm}

For imbalanced data, the WSVM can easily be adopted as the base classifier for multilevel framework (MLWSVM). The regular SVM does not  perform well on imbalanced data because it tends to train models with respect to the majority class and technically ignores the minority class. However, the effect of imbalanced issue decreases while using multilevel framework since we prevent creating very small coarse sets for the minority class even if the majority class can still be coarsened.

Often, methods for imbalanced classification demonstrate poor performance on data with missing values (such as \cite{farhangfar2008impact}) that is a frequent situation in healthcare data. Therefore, we apply imputation methods prior the classification model. Such imputation methods have been well studied in  statistical analysis and machine learning domains \cite{ghannad2010selection,little2002statistical,schafer2010analysis,garcia2010pattern,gheyas2010neural}. Problems with missing data can be categorized into three types: data is completely at random (MCAR), missing at random (MAR), and not missing at random (NMAR). MCAR occurs while any feature of a data instance is missing completely random and is independent of the values of other features. Data is MAR , when the data instance with missing feature is dependent on the value of one or more of the instances’ other features. NMAR occurs when the data instance with missing feature is dependent on the value of the other missing features. Even though MCAR is more desirable, in many real-world problems, MAR occurs frequently in practice \cite{ghannad2010selection}.


In the imputation methods, the goal is to substitute a missing value with a meaningful estimation \cite{garcia2010pattern}. This can be done  either directly from the information on the dataset or by constructing a predictive model for this purpose. Standard methods for imputation are mean imputation \cite{donders2006review}, kNN imputation \cite{batista2002study}, Bayesian principal component analysis (BPCA) imputation \cite{oba2003bayesian}, and the expectation maximization (EM) \cite{schneider2001analysis}. We apply the EM method which is  one of the most successful imputation methods \cite{huang2009maximum}. The EM method  iteratively applies linear regression analysis and fits a new linear to the estimated data until a local optimum is achieved \cite{ghahramani1994supervised,schneider2001analysis}.  In the regularized adaption of EM method, the conditional maximum likelihood estimation of regression parameters is replaced in the conventional EM algorithm \cite{nanni2012classifier}.

\subsection{Regularized Expectation-Maximization}
In our preprocessing, when the data contains many missing values, we apply the EM algorithm. It iteratively calculates the maximum-likelihood (ML) estimates of parameters by exploring the relationship between the complete  and incomplete data (with missing features) \cite{dempster1977maximum}. In many cases, it has been demonstrated that the EM algorithm achieves a reliable global convergence to a local maximizer (from almost any starting point), and economical storage. It is not computationally expensive, and can be easily implemented \cite{redner1984mixture}. The EM algorithm maximizes the log-likelihood ($L$) of the incomplete data
\begin{subequations}
\begin{align}
              L(\Theta;\chi) = \sum_{i=1}^{n}\log p(x_i| \Theta),
\end{align}
\end{subequations}
where $\chi = \{x_i| i=1,...,n\}$ are the observations with independent distribution $p(x)$ parameterized by $\Theta$ and P is the distribution function of the complete data given $\Theta$.
The regularized EM algorithm (REM) is developed to control the level of uncertainty associated to missing values \cite{li2005regularized}. The main idea is to regularize the likelihood function according to the mutual relationship between the observations and the missing data with little uncertainty and maximum information. Intuitively, it is desirable to select the missing data that has a high probabilistic association with the observations, which shows that there is little uncertainty on the missing data given the observations. It performs linear regression iteratively for the imputation of missing values. The REM algorithm optimizes the penalized likelihood as follows:
\begin{subequations}
\begin{align}
             \tilde{L}(\Theta;\chi) = L(\Theta;\chi) + \Gamma P(\chi,\Upsilon| \Theta).
\end{align}
\end{subequations}
The trade-off between the degree of regularization of the solution and the likelihood function is controlled by the so-called regularization parameter that is represented by $\Gamma$  \cite{li2005regularized}.
In addition to reducing the uncertainty of missing data, the REM preserves the advantage of the standard EM method. This method is very efficient for over-complicated models.

The EM algorithm implementation in this paper is based on iterated linear regression analysis. In the regularized EM algorithm, a regularized estimation method substitutes the conditional maximum likelihood estimation of regression parameters in the conventional EM algorithm. We used the modules from  \cite{schneider2001analysis}, which apply truncated total least squares (with fixed truncation parameter) and ridge regression with generalized cross-validation as regularized estimation methods. We only perform the REM imputation for all classification datasets in the paper.

\subsection{Performance Measures}
Classification algorithms are evaluated using the performance measures calculated from the confusion matrix (see Table \ref{wrap_tab_1}).
\begin{table}[!ht]
\centering
\caption{Confusion matrix}\label{wrap_tab_1}
\begin{tabular}{|c|c|c|}
\hline & Positive class & Negative Class\\  \hline
\multirow{2}{*}{Positive Class} & True Positive & False Positive   \\
                                  &   (TP)         &     (FP)          \\    \hline
 \multirow{2}{*}{Negative Class} & False Negative & True Negative  \\
                                &   (FN)            &    (TN)          \\
\hline
\end{tabular}
\end{table}

For binary classification problems, the performance measures are defined as sensitivity (SN), specificity (SP), and G-mean, accuracy (ACC), namely,
\begin{equation}
\textrm{SN}=\frac{TP}{TP+FN}, \ \ \textrm{SP}=\frac{TN}{TN+FP},
\end{equation}
\begin{equation}
\textrm{G-mean}=\sqrt{\textrm{SP}*\textrm{SN}},
\end{equation}
and
\begin{equation}
 \textrm{ACC}=\frac{TP+TN}{FP+TN+TP+FN}.
\end{equation}

\section{Numerical results} \label{sec:numres}
Due to proprietary nature of medical data, anonymized medical records can be made available for research purposes, but cannot be shared in open access. For validation and reproduction, it is helpful to first examine the performance of proposed methods on standard public data sets. Our source code is available at \cite{plos-svm-impl}.

We evaluate the proposed classification framework on public (UCI \cite{UCI}, and the cod-rna dataset \cite{alon1999broad}), and healthcare proprietary binary classification benchmarks \cite{RoderickMEDAI}, \cite{RoderickFLU}. Both the coarsest and refinement (W)SVM models are solved using LIBSVM-3.18 \cite{chang2011libsvm}, and the FLANN library \cite{muja_flann_2009} is used to create the A$k$NN graphs.
We used a 'composite' algorithm of FLANN, which is a combination of multiple randomized KD trees and hierarchical k-means trees. According to \cite{muja_flann_2009}, it outperforms separate KD trees and hierarchical k-means trees, so we report only the results of the `composite' algorithm in FLANN. The number of requested nearest neighbors in A$k$NN is selected as $k=10$. Increasing $k$ does not improve the results. We chose $P$ for the number of nearest opposite-class clusters as 10\% of the number of clusters in the corresponding class (Algorithm \ref{alg3}, line 16). 
In particular, after clustering each class individually, instead of training each cluster in one class with all clusters of the other class, we will pick 10\% of clusters that belong to the other class, in a way that these clusters are the closest clusters to the current cluster $k$. Next, we train cluster $k$  with these 10\% of nearest opposite clusters. For example, if there are 20 clusters in the majority class and 5 clusters in the minority class, we pick a cluster in the minority class and train it with  2  clusters of the majority class. 
Multilevel frameworks, data processing and further scripting are implemented in MATLAB 2012a  \cite{MATLAB:2012}.
The C4.5 \cite{quinlan1986induction}, Naive Bayes (NB) \cite{russell2002artificial}, Logistic Regression (LR) \cite{homser1989applied}, and 5-Nearest Neighbor (5NN) \cite{cover1967nearest} are implemented using WEKA \cite{hall2009weka} interfaced with MATLAB. A typical 10-fold cross validation setup is used.
We create missing values on the public data training sets by discarding the features randomly. The misclassification penalty or weights are selected as inversely proportional to the size of each class in our implementation. As a preprocessing step, the whole data is normalized such that it has a zero mean and unitary standard deviation. before classification. The nested uniform design (UD) is performed on the training data as the model selection for (W)SVM \cite{huang2007model}. The UD methodology is very successful for model selection in supervised learning \cite{mangasarian2008privacy}. The close-to-optimal parameter set is achieved in an iterative nested process \cite{huang2007model}. The optimal parameter set is selected based on {\em G-mean} maximization, since data might be imbalanced. A 9- and 5-point run design is performed for the first and second stages of the nested UD due to its superiority for the UCI data \cite{huang2007model}, and the performance measures such as sensitivity, specificity, G-mean and accuracy are calculated on the testing data.

The nested UD method is performed in two stages. In the first stage, a
9-run UD sampling pattern is conducted in the appropriate search range for C and $\gamma$. In the second stage, the search range for each parameter is fixed around the best point from the first stage. Then a 5-runs UD
sampling pattern is searched in the new range. The total number of parameter combinations
is 13 (the center point at the second stage is the duplicate point which is
trained and should be considered only once). The optimal parameters are determined inside the MLSVM model each time before SVM training each time. The initial range of parameter $C$ is between 0.01 and 100, the initial range of parameter $\gamma$ is between 0.005000 and 3.000078 for the nested UD. For the REM implementation, we used multiple ridge regression within 5-fold cross-validation. We performed the REM imputation in each fold of cross validation on the training data (90\% of the whole set). This means that there will be no transfer of information from the validation data set into the training data through the imputation scheme to avoid biased results.

\begin{table}[h]
\centering
\caption{Public data sets.}
\label{table1}
    \begin{tabular}{cccccccc}         

        \hline
Dataset	&	$r_{imb}$	&	$n_f$ &	$|\JJ|$	&	$|\CC^+|$	&	$|\CC^-|$	\\ \hline
Twonorm	&	0.50	&	20	&	7400	&	3703	&	3697	\\
Letter26	&	0.96	&	16	&	20000	&	734	&	19266	\\
Ringnorm	&	0.50	&	20	&	7400	&	3664	&	3736	\\
Cod-rna	&	0.67	&	8	&	59535	&	19845	&	39690	\\
Clean (Musk)	&	0.85	&	166	&	6598	&	1017	&	5581	\\
Advertisement	&	0.86	&	1558	&	3279	&	459	&	2820	\\
Nursery	&	0.67	&	8	&	12960	&	4320	&	8640	\\
Hypothyroid	&	0.94	&	21	&	3919	&	240	&	3679	\\
Buzz  &	0.80	&	77	&	140707	&	27775	&	112932	\\
Forest   & 0.98   &   54 & 581012   & 9493   &  571519      \\
\hline
\end{tabular}
\end{table}

\subsection{Public data sets}
We compared several methods with the proposed ML(W)SVM to classify data with missing values.
In Table \ref{table2} we show the comparative results of MLSVM, MLWSVM, SVM, WSVM, Naive Bayes, C4.5, LR, and 5NN algorithms  evaluated on public data sets in Table \ref{table1}. 
These methods are examined for different missing value ratios ($r_{mv}$) selected as 5\%, 10\%, 20\%, and 40\%. We used the REM method for missing data imputation \cite{schneider2001analysis}. The best results for their corresponding missing values' levels among all methods are shown in bold which makes clear that MLWSVM and WSVM perform better than the other methods in general for all missing value ratios (see last row in Table \ref{table2}). In fact, MLWSVM and WSVM result in higher G-mean values in 22 out of 40  dataset-$r_{mv}$ combinations followed by MLSVM, SVM, and C4.5 with 13 out of 40. 
Moreover, the ML(W)SVM techniques achieve faster computational time in comparison to the standard (W)SVM (Table \ref{ta2}).
We note that, while the results on publicly available data are easy to relate to and reproduce, testing on relatively small and well-known data sets has its limitations. Most standard methods and implementations achieve acceptable metrics of quality because have been tested on and tuned using these data sets. In practice, we do not expect that a specialized method will show a dramatic, consistent improvement in quality. A fast  computational time without any loss in quality is the most significant result in this work, which illustrates the advantage of multi-level approach that inherits flexible SVM parameters found at coarser levels.

\begin{table*}[h]
\footnotesize
\centering
\caption{Comparative G-mean results for ML(W)SVM against the regular SVM, WSVM, NB, C4.5, 5NN, and LR on academic datasets for different fractions of missing values ($r_{mv}$) using the REM imputation method.}
\label{table2}
    \begin{tabular}{cccccccccc}                  
        \hline
Dataset	&	$r_{mv}$	&	MLSVM	&	MLWSVM	&	SVM	&	WSVM	&	C4.5	&	5NN	&	NB	&	LR	\\ \hline

\multirow{4}{*}{Twonorm}	&	5\%	&	{\bf 0.98}	&	{\bf 0.98}	&	{\bf 0.98}	&	{\bf 0.98}	&	0.86	&	0.97	&	{\bf 0.98}	&	{\bf 0.98}	\\
	&	10\%	&	{\bf 0.98}	&	{\bf 0.98}	&	0.97	&	0.97	&	0.87	&	0.97	&	0.97	&	0.97	\\
	&	20\%	&	{\bf 0.98}	&	{\bf 0.98}	&	{\bf 0.98}	&	{\bf 0.98}	&	0.88	&	0.97	&	0.97	&	{\bf 0.98}	\\
	&	40\%	&	0.97	&	0.97	&	0.97	&	0.97	&	0.89	&	0.97	&	{\bf 0.98}	&	{\bf 0.98}	\\\hline
\multirow{4}{*}{Letter}	&	5\%	&	0.97	&	{\bf 1.00}	&	0.99	&	0.99	&	0.97	&	0.98	&	0.86	&	0.81	\\
	&	10\%	&	0.98	&	{\bf 1.00}	&	0.98	&	0.99	&	0.98	&	0.98	&	0.86	&	0.80	\\
	&	20\%	&	{\bf 1.00}	&	{\bf 1.00}	&	0.99	&	0.99	&	0.97	&	0.98	&	0.87	&	0.80	\\
	&	40\%	&	0.95	&	0.97	&	0.96	&	{\bf 0.99}	&	0.97	&	0.98	&	0.88	&	0.83	\\\hline
\multirow{4}{*}{Ringorm}	&	5\%	&	0.97	&	0.98	&	0.97	&	0.98	&	0.91	&	0.61	&	{\bf 0.99}	&	0.76	\\
	&	10\%	&	0.98	&	0.98	&	{\bf 0.99}	&	{\bf 0.99}	&	0.91	&	0.62	&	0.98	&	0.76	\\
	&	20\%	&	{\bf 0.98}	&	{\bf 0.98}	&	0.97	&	{\bf 0.98}	&	0.91	&	0.62	&	{\bf 0.98}	&	0.76	\\
	&	40\%	&	{\bf 0.98}	&	{\bf 0.98}	&	0.97	&	{\bf 0.98}	&	0.91	&	0.62	&	{\bf 0.98}	&	0.76	\\\hline
\multirow{4}{*}{Cod-rna}	&	5\%	&	0.95	&	{\bf 0.96}	&	{\bf 0.96}	&	{\bf 0.96}	&	0.95	&	0.92	&	0.66	&	0.93	\\
	&	10\%	&	0.95	&	{\bf 0.96}	&	0.95	&	{\bf 0.96}	&	0.95	&	0.91	&	0.66	&	0.92	\\
	&	20\%	&	0.95	&	{\bf 0.96}	&	0.95	&	0.95	&	0.94	&	0.91	&	0.67	&	0.92	\\
	&	40\%	&	{\bf 0.95}	&	{\bf 0.95}	&	{\bf 0.95}	&	{\bf 0.95}	&	0.93	&	0.90	&	0.68	&	0.91	\\\hline
\multirow{4}{*}{Clean}	&	5\%	&	{\bf 1.00}	&	0.99	&	0.98	&	{\bf 1.00}	&	0.83	&	0.92	&	0.79	&	0.89	\\
	&	10\%	&	0.99	&	{\bf 1.00}	&	0.99	&	{\bf 1.00}	&	0.83	&	0.91	&	0.79	&	0.89	\\
	&	20\%	&	{\bf 1.00}	&	{\bf 1.00}	&	{\bf 1.00}	&	{\bf 1.00}	&	0.83	&	0.91	&	0.79	&	0.89	\\
	&	40\%	&	{\bf 1.00}	&	{\bf 1.00}	&	{\bf 1.00}	&	{\bf 1.00}	&	0.82	&	0.92	&	0.79	&	0.89	\\\hline
\multirow{4}{*}{Advertisement}	&	5\%	&	0.87	&	0.87	&	0.87	&	0.87	&	{\bf 0.92}	&	0.81	&	0.60	&	0.82	\\
	&	10\%	&	{\bf 0.87}	&	{\bf 0.87}	&	0.86	&	0.86	&	0.86	&	0.85	&	0.62	&	0.82	\\
	&	20\%	&	0.83	&	0.85	&	0.83	&	0.85	&	{\bf 0.89}	&	0.83	&	0.61	&	0.83	\\
	&	40\%	&	0.84	&	0.86	&	0.87	&	0.81	&	{\bf 0.91}	&	0.85	&	0.62	&	0.82	\\\hline
\multirow{4}{*}{Nursery}	&	5\%	&	0.99	&	0.99	&	{\bf 1.00}	&	{\bf 1.00}	&	{\bf 1.00}	&	{\bf 1.00}	&	0.00	&	{\bf 1.00}	\\
	&	10\%	&	0.99	&	0.99	&	{\bf 1.00}	&	{\bf 1.00}	&	{\bf 1.00}	&	{\bf 1.00}	&	0.00	&	{\bf 1.00}	\\
	&	20\%	&	0.96	&	0.96	&	{\bf 1.00}	&	{\bf 1.00}	&	{\bf 1.00}	&	{\bf 1.00}	&	0.00	&	{\bf 1.00}	\\
	&	40\%	&	0.92	&	0.92	&	{\bf 1.00}	&	{\bf 1.00}	&	{\bf 1.00}	&	0.99	&	0.46	&	{\bf 1.00}	\\\hline
\multirow{4}{*}{Hypothyroid}	&	5\%	&	0.83	&	0.87	&	0.81	&	0.87	&	0.96	&	0.76	&	{\bf 0.97}	&	0.88	\\
	&	10\%	&	0.85	&	0.86	&	0.78	&	0.86	&	{\bf 0.96}	&	0.76	&	{\bf 0.96}	&	0.89	\\
	&	20\%	&	0.84	&	0.86	&	0.72	&	0.86	&	0.96	&	0.75	&	{\bf 0.97}	&	0.90	\\
	&	40\%	&	0.86	&	0.88	&	0.84	&	0.88	&	0.96	&	0.76	&	{\bf 0.97}	&	0.89	\\\hline
\multirow{4}{*}{Buzz}	&	5\%	&	{\bf 0.94}	&	{\bf 0.94}	&	{\bf 0.94}	&	{\bf 0.94}	&	{\bf 0.94}	&	0.93	&	0.89	&	{\bf 0.94}	\\
	&	10\%	&	{\bf 0.94}	&	{\bf 0.94}	&	{\bf 0.94}	&	{\bf 0.94}	&	{\bf 0.94}	&	0.93	&	0.89	&	{\bf 0.94}	\\
	&	20\%	&	0.92	&	{\bf 0.94}	&	0.93	&	{\bf 0.94}	&	{\bf 0.94}	&	0.93	&	0.88	&	0.93	\\
	&	40\%	&	0.93	&	0.93	&	0.93	&	0.93	&	{\bf 0.94}	&	{\bf 0.94}	&	0.86	&	{\bf 0.94}	\\
\hline
\multirow{4}{*}{Forest}	&	5\%	&	0.90	&	{\bf 0.91}	&	0.90	&	{\bf 0.91}	&	{\bf 0.91}	&	0.87	&	0.80	&	0.00	\\
&	10\%	&	0.92	&	{\bf 0.93}	&	0.91	&	0.92	&	0.88	&	0.85	&	0.78	&	0.00	\\
&	20\%	&	0.91	&	{\bf 0.92}	&	0.90	&	0.91	&	0.89	&	0.84	&	0.77	&	0.00	\\
&	40\%	&	0.88	&	0.89	&	0.88	&	{\bf 0.90}	&	0.85	&	0.82	&	0.73	&	0.00    \\
\hline
$\#$ of bold values & & 13  & 22	&13	&	22 &13	&	4 & 9	&	10	   \\
\hline
\end{tabular}
\end{table*}
\normalsize

\begin{table}[h]
\footnotesize
\centering
\caption{Computational time in seconds (not including the REM method)}
\label{ta2}
    \begin{tabular}{ccccc}                  
        \hline
	&	MLSVM	&	SVM	&	MLWSVM	&	WSVM	\\\hline
Twonorm	&	5	&	28	&	5	&	28	\\
Letter	&	30	&	138	&	32	&	139	\\
Ringnorm	&	4	&	25	&	4	&	26	\\
Cod-rna	&	266	&	1831	&	281	&	1857	\\
Clean	&	17	&	95	&	15	&	82	\\
Advertisement	&	98	&	227	&	100	&	231	\\
Nursery	&	25	&	187	&	31	&	192	\\
Hypothyroid	&	2	&	3	&	2	&	3	\\
Buzz	&	2209	&	25257	&	2999	&	26026	\\
Forest  &     13328      &  352500         &    13360       &    353210       \\
\hline
\end{tabular}
\end{table}

%

\subsection{Healthcare data sets}
We present the results of comparison of classification algorithms on the real-life healthcare data sets.
In Table \ref{table3} we show the results on Example 1 (see Section \ref{sec:intro}, and Table  \ref{healthtable}), a classification task of assigning a patient in a correct group by financial risk, which are ordered in ascending manner from group 1 with the lowest level of risk, to group 5 with the highest level of risk. The data used in the study was provided by Geisinger Health System in a follow-up study to the internal report on prediction on integrated clinical and financial data  \cite{RoderickMEDAI}.

\begin{table}[h]
\caption{Healthcare datasets. The set ``Example 1'' has 10000 observations in each class. In set ``Example 2'', the majority and minority classes contain 50400, and 33600 observations, respectively. For details about the data see \cite{RoderickFLU}.}
\label{healthtable}
\centering
\begin{tabular}{cccc}
	\hline Data  &  $n_f$ &	$|\JJ|$ & No. of classes  \\ \hline
	 Example 1 & 16 & 50000 & 5\\
	 Example 2 & 13 & 84000 & 2\\
        	\hline
\end{tabular}
\end{table}


The motivation behind the original study was to determine how much integration of the medical and financial data changes the outcomes of clustering and classification operations based on financial data alone. In the original study, a logistic regression (LR) (implemented as \textit{mnrfit} in MATLAB) was used as a default binary classifier; other standard choices such as nearest-neighbor or naive Bayesian classifiers were rejected in the original report as producing lower quality of prediction. Our comparison illustrates the point that a specialized method developed for imbalanced, incomplete data here outperforms an approach that is accepted as default for a healthcare application.

 We compare the accuracy of commonly used in healthcare data analysis LR with that obtained by ML(W)SVM. The strategy ``one-against-all`` is used for multi-class classification. This strategy performs training a classifier per class with the data points of that class as positive class and the rest of the data points are trained as negative class. Results in Tables \ref{table3}-\ref{table4} have also been obtained using 10-fold cross validation.


\begin{table}[h]
\caption{Accuracy of financial risk problem with five risk classes (Example 1) using the REM imputation method.}
\label{table3}
\centering
\begin{tabular}{cccccc}
	\hline Class  & 1 & 2 & 3 & 4 & 5 \\ \hline
	 LR & 0.58 & 0.54 & 0.53 & 0.51 & 0.59 \\
	 MLSVM & 0.83 & 0.78 & 0.77 & 0.78 & 0.90\\
           MLWSVM & 0.86&0.76&0.76& 0.77&0.91 \\
	\hline
\end{tabular}
\end{table}


\begin{table}[h]
\caption{Sensitivity, specificity and G-mean of financial risk problem with five risk classes (Example 1) using ML(W)SVM and REM imputation methods.}
\label{Addded_table}
\centering
\begin{tabular}{ccccccc}
        \hline &  \multicolumn{3}{c}{MultilevelSVM}  & \multicolumn{3}{c}{MultilevelWSVM} \\
	\hline   & SN & SP & G-mean  & SN & SP & G-mean \\ \hline
Class 1 &	0.86&	0.73&	0.79&	0.89&	0.74&	0.81	\\
Class 2 &	0.89&	0.34&	0.55&	0.86&	0.36&	0.56	\\
Class 3 &	0.89&	0.28&	0.50&	0.88&	0.29&	0.50	\\
Class 4 &	0.88&	0.40&	0.60&	0.87&	0.40&	0.58	\\
Class 5 &	0.96&	0.69&	0.81&	0.96&	0.70&	0.82	\\
	\hline
\end{tabular}
\end{table}

To interpret the results, we note that correct identification of intermediate risk categories is a very difficult problem in medical informatics. To our knowledge, there is no good definition of ``average health``, either evidence-driven or philosophical, that would help an expert to identify such patient features that do not indicate an acute crisis, or an almost certain safety from crisis. Accordingly, it is not surprising that neither approach does well on the risk categories 2-4; there is also not a lot of motivation to improve the model there.
On the other hand, it is important to identify and predict the very low-risk patients (knowing that status ahead of time allows resource re-allocation leading to savings and improved service for everyone) and the very high-risk patients (so that clinical and financial resources could be prepared for the forthcoming crisis).

Accordingly, it is important that the use of an advanced method of machine learning changes the quality of prediction from almost worthless ('toss a coin') to workable (accuracy of $0.7$).

In Table \ref{table4}, we compare results for the widely used basic approach and ML(W)SVM prediction for Example 2 (see Section \ref{sec:intro}), a study of patient's response to hospital flu outreach. In this problem, the goal is to find a binary classifier that will predict whether the patient will get vaccinated after reminder, or not (this includes using a different provider for vaccination). In the preliminary study, we used adaptive linear regression model (LASSO for adaptive selection of features, logistic regression on actual prediction).

\begin{table}[h]
\caption {Comparison of Multilevel WSVM against Multilevel SVM and Adaptive Logistic Regression (LR)  using the REM imputation method. Improved results are in bold.}
\label{table4}
\centering
\begin{tabular}{ccccc}
	\hline &G-mean & SN   & SP & ACC \\
	\hline Adaptive LR& 0.7516  & 0.8903 & 0.6345 & 0.7619 \\
	 MLSVM & {\bf 0.8012}   & {\bf 0.9750} & {\bf 0.6583} & {\bf 0.8496} \\
	 MLWSVM & {\bf 0.8016}     & {\bf 0.9739} & {\bf 0.6598} & {\bf 0.8495} \\
	\hline
\end{tabular}
\end{table}

Response to outreach is not a crucial life-or-death issue, we are performing this study to see if predictive modeling can assist with resource allocation (which patients to contact, how much medical personnel effort to dedicate to outreach and then vaccination). Arguably, accuracy is more important than specificity here.
Even the basic results (using linear regression) were met with approval the CPSL (Care Patient Service Line: a division responsible for coordinating efforts of local, small-scale healthcare providers operating under Geisinger). SVM methods (almost 10 percent improvement) provide additional justification for the use of machine learning on merged data to assist planning in clinical practice.
%
\section{Discussion}

Large-scale data, missing or imperfect features, skewness distribution of classes are common challenges in pattern recognition of many healthcare problems. We have successfully extended a powerful machine learning technique, support vector machines, to the \emph{scalable multilevel framework} of cost-sensitive learning SVM to deal with imbalanced classification problems. Our multilevel framework substantially improves the computational time without losing the quality of classifiers for large-scale datasets. We have shown that MLWSVM produces superior results than MLSVM and the regular SVM methods in most cases. This work can be extended to tackle other classification problems with large-scale imbalanced data (combined from different sources) with missing features in healthcare and engineering applications.

From the perspective of evidence-driven healthcare, our work shows that application of cutting edge machine learning techniques (in this case, fast multilevel classifiers) makes enough of a difference to justify the additional development effort for typical examples from clinical practice. While the improvements in precision and specificity we show in this study are both under $10 \%$ and are modest in general perspective, the result in healthcare is significant.

To our knowledge, such complex combined behavioral/operational phenomena as inference of financial risk from medical history (Example 1), or prediction of effectiveness of public outreach (Example 2), don't have a satisfactory casual explanation. The classical (1990s) clinical practice offered two equally unsatisfactory options: not having a capability for prediction at all, or relying on very basic statistical techniques (based on a single data source, with very high rate of false-positive classification outcomes). The existing mature models (such as actuarial projections of financial risk) do not benefit from integration of data from multiple sources, and may, in fact, turn out to be ineffective outside of their scope in patient population and metrics of interest (as we have shown in \cite{RoderickMEDAI}). Thus, in the modern clinical practice we have to rely on newly developed machine learning tools, tuned on data from multiple sources.
Thus, our work can also be extended to handle other classification problems on massive, multi-format medical data.

The long-term healthcare impact of this type of work consists of two parts: to demonstrate general advantages of applying a specialized machine learning approach to healthcare data, and to argue for the use of multi-scale representation on complex medical data, integrated from multiple sources and containing rare events. Although the results presented here are not ideal (possibly due to complexity of the studied phenomena), they are sufficient to recommend the method for future use in healthcare predictive analytics.

\section*{Data sharing policy}
Clinical data and aggregated medical insurance data in this study were provided, correspondingly, by Geisinger Health System and Geisinger Health Plan. Due to considerations of patient privacy, the data is not available for free access. However, it can be shared, in de-identified form, for research purposes. To access these data sets, please contact the corresponding author.
\nolinenumbers

%
%
%
\bibliographystyle{plos2015}
\bibliography{roderick_medical,paper,ilya}

\newpage

\section*{Supplementary Materials}

{\bf Details about the REM method}

\begin{itemize}
   \item The parameter P and $\theta$ are calculated based on Shannon information theory \cite{li2005regularized}.
   \item To implement this algorithm, we refer the reader to \cite{Schneider2001} in which the implementation is provided. The EM algorithm in this package is based on iterated linear regression analyses. In the regularized EM algorithm, a regularized estimation method substitutes the conditional maximum likelihood estimation of regression parameters in the conventional EM algorithm. The modules in  \cite{Schneider2001} present the truncated total least squares (with fixed truncation parameter) and ridge regression with generalized cross-validation as regularized estimation methods. In our application, we used multiple ridge regression, stagnation tolerance parameter = 1e-2 (stop criteria for iteration), maximum number of EM iterations = 50, truncation parameter selection: 'KCV' (that chooses a truncation adaptively for each record by K-fold cross-validation), parameter K = 5 for K-fold cross-validation, norm of error to be minimized in K-fold cross-validation = 2.
\end{itemize}


{\bf  Description of the cross validation scheme}
\begin{itemize}
   \item We performed 10-cross validation on the data. For cross validation purposes, 90\% of the data was used for training and the rest 10\% was used for testing. The data is normalized prior to classification, so that it has zero mean and unitary standard deviation. At each fold, first the training set is coarsened level by level until we reach the coarsest level and then the parameter optimization (UD) is applied to find the optimal $C$, $C^+$, $C^-$, and $\gamma$. Next, we train (W)SVM and get the support vectors. Then, we update the training data based on the support vectors (SVs) of the coarse level with points in the fine level that are close to SVs. Finally, we apply the UD on the updated training data with setting of optimal $C$, $C^+$, $C^-$ and $\gamma$ of previous level as corresponding initial paramenters for the next finer level training. We will continue till we reach the finest level. At the finest level we calculate the performance measure for that fold and continue to run the same V-cycle model (MLWSVM) ten times for each training and testing set that are selected randomly each time to make sure that all parts of the data are considered. Finally, we report the average performance measures over 10-fold cross validation.
\end{itemize}

{\bf  MLWSVM framework}
\begin{itemize}
   \item The source code will be available at https://people.cs.clemson.edu/~isafro/software.html after final acceptance of the paper.
\end{itemize}

{\bf  Supporting tables}
\begin{itemize}
   \item Tables \ref{tableadded1},\ref{tableadded2},\ref{tableadded3} show the sensitivity, specificity, and accuracy of comparative algorithms on public data sets.
    \item Table \ref{tableadded4} shows the computational time for C4.5, 5NN, NB, LR, and MLSVM (excluding model selection) on public datasets.
\end{itemize}

\begin{table*}[h]
\footnotesize
\centering
\caption{Comparative sensitivity results for ML(W)SVM against the regular SVM, WSVM, NB, C4.5, 5NN, and LR on Twonorm, Letter, Ringnorm, and Clean academic datasets for different fractions of missing values ($r_{mv}$) using the REM imputation method.}
\label{tableadded1}
    \begin{tabular}{cccccccccc}                  
        \hline
Dataset	&	$r_{mv}$	&	MLSVM	&	MLWSVM	&	SVM	&	WSVM	&	C4.5	&	5NN	&	NB	&	LR	\\ \hline
\multirow{4}{*}{Twonorm}	&	5\%	&	0.97	&	0.98	&	0.97	&	0.98	&	0.87	&	0.97	&	0.97	&	0.97	\\
	&	10\%	&	0.98	&	0.98	&	0.97	&	0.97	&	0.85	&	0.96	&	0.97	&	0.97	\\
	&	20\%	&	0.99	&	0.99	&	0.99	&	0.99	&	0.86	&	0.97	&	0.98	&	0.98	\\
	&	40\%	&	0.96	&	0.97	&	0.97	&	0.97	&	0.88	&	0.97	&	0.96	&	0.97	\\\hline
\multirow{4}{*}{Letter}	&	5\%	&	1.00	&	1.00	&	1.00	&	1.00	&	1.00	&	0.99	&	0.96	&	0.97	\\
	&	10\%	&	1.00	&	1.00	&	1.00	&	1.00	&	1.00	&	1.00	&	0.97	&	1.00	\\
	&	20\%	&	0.99	&	0.99	&	0.99	&	0.99	&	1.00	&	1.00	&	0.97	&	1.00	\\
	&	40\%	&	1.00	&	1.00	&	1.00	&	1.00	&	1.00	&	1.00	&	0.97	&	0.99	\\\hline
\multirow{4}{*}{Ringnorm}	&	5\%	&	0.97	&	0.98	&	0.97	&	0.98	&	0.90	&	1.00	&	0.99	&	0.81	\\
	&	10\%	&	0.96	&	0.96	&	0.96	S&	0.96	&	0.91	&	1.00	&	0.99	&	0.81	\\
	&	20\%	&	0.98	&	0.98	&	0.98	&	0.98	&	0.91	&	1.00	&	0.99	&	0.81	\\
	&	40\%	&	0.96	&	0.98	&	0.98	&	0.98	&	0.90	&	1.00	&	0.99	&	0.81	\\\hline
\multirow{4}{*}{Clean}	&	5\%	&	1.00	&	0.99	&	0.99	&	0.99	&	0.89	&	0.99	&	0.86	&	0.99	\\
	&	10\%	&	0.99	&	0.99	&	0.99	&	1.00	&	0.86	&	0.99	&	0.86	&	0.98	\\
	&	20\%	&	1.00	&	1.00	&	1.00	&	1.00	&	0.80	&	1.00	&	0.90	&	0.99	\\
	&	40\%	&	1.00	&	1.00	&	1.00	&	1.00	&	0.95	&	0.99	&	0.87	&	0.98	\\
\hline
\multirow{4}{*}{Forset}	&	5\%	&0.98	&0.97	&0.99	&0.99	&1.00	&1.00	&0.99	&1.00\\
&10\%&	0.96&	0.98&	0.97&	0.98&	1.00&	1.00&	0.99&	1.00
\\
&20\%&	0.98&	0.97&	0.96&	0.97&	1.00&	1.00&	0.99&	1.00\\
&40\%&	0.98&	0.96&	0.97&	0.97&	1.00&	1.00&	0.99&	1.00\\

\hline
\end{tabular}
\end{table*}
\normalsize

\begin{table*}[h]
\footnotesize
\centering
\caption{Comparative specificity results for ML(W)SVM against the regular SVM, WSVM, NB, C4.5, 5NN, and LR on Twonorm, Letter, Ringnorm, and Clean academic datasets for different fractions of missing values ($r_{mv}$) using the REM imputation method.}
\label{tableadded2}
    \begin{tabular}{cccccccccc}                  
        \hline
Dataset	&	$r_{mv}$	&	MLSVM	&	MLWSVM	&	SVM	&	WSVM	&	C4.5	&	5NN	&	NB	&	LR	\\ \hline
\multirow{4}{*}{Twonorm}	&	5\%	&	0.98	&	0.97	&	0.98	&	0.97	&	0.87	&	0.97	&	0.98	&	0.98	\\
	&	10\%	&	0.99	&	0.98	&	0.97	&	0.97	&	0.86	&	0.97	&	0.98	&	0.97	\\
	&	20\%	&	0.97	&	0.97	&	0.97	&	0.97	&	0.87	&	0.97	&	0.98	&	0.98	\\
	&	40\%	&	0.98	&	0.99	&	0.98	&	0.99	&	0.88	&	0.97	&	0.98	&	0.98	\\\hline
\multirow{4}{*}{Letter}	&	5\%	&	0.95	&	0.99	&	0.98	&	0.98	&	0.95	&	0.98	&	0.77	&	0.68	\\
	&	10\%	&	0.96	&	0.99	&	0.97	&	0.98	&	0.97	&	0.95	&	0.77	&	0.65	\\
	&	20\%	&	1.00	&	1.00	&	1.00	&	1.00	&	0.94	&	0.96	&	0.80	&	0.64	\\
	&	40\%	&	0.90	&	0.94	&	0.91	&	0.94	&	0.95	&	0.97	&	0.79	&	0.69	\\\hline
\multirow{4}{*}{Ringnorm}	&	5\%	&	0.98	&	0.99	&	0.98	&	0.99	&	0.91	&	0.38	&	0.98	&	0.72	\\
	&	10\%	&	1.00	&	1.00	&	1.00	&	1.00	&	0.91	&	0.38	&	0.98	&	0.72	\\
	&	20\%	&	0.98	&	0.99	&	0.98	&	0.99	&	0.91	&	0.37	&	0.98	&	0.71	\\
	&	40\%	&	1.00	&	0.99	&	0.98	&	0.99	&	0.90	&	0.37	&	0.98	&	0.71	\\\hline
\multirow{4}{*}{Clean}	&	5\%	&	0.99	&	0.99	&	0.98	&	1.00	&	0.78	&	0.85	&	0.74	&	0.80	\\
	&	10\%	&	0.98	&	1.00	&	0.99	&	1.00	&	0.81	&	0.84	&	0.73	&	0.81	\\
	&	20\%	&	0.98	&	0.99	&	0.99	&	1.00	&	0.86	&	0.82	&	0.70	&	0.80	\\
	&	40\%	&	1.00	&	1.00	&	0.99	&	1.00	&	0.71	&	0.85	&	0.72	&	0.81	\\
\hline
\multirow{4}{*}{Forset}	
&5\%	&	0.84	&	0.84	&	0.83	&	0.84	&	0.82	&	0.76	&	0.65	&	0.00	\\
&10\%	&	0.87	&	0.87	&	0.86	&	0.87	&	0.78	&	0.73	&	0.61	&	0.00	\\
&20\%	&	0.85	&	0.87	&	0.84	&	0.85	&	0.79	&	0.71	&	0.60	&	0.00	\\
&40\%	&	0.79	&	0.84	&	0.80	&	0.83	&	0.73	&	0.67	&	0.54	&	0.00	\\
\hline
\end{tabular}
\end{table*}
\normalsize

\begin{table*}[h]
\footnotesize
\centering
\caption{Comparative accuracy results for ML(W)SVM against the regular SVM, WSVM, NB, C4.5, 5NN, and LR on Twonorm, Letter, Ringnorm, and Clean academic datasets for different fractions of missing values ($r_{mv}$) using the REM imputation method.}
\label{tableadded3}
    \begin{tabular}{cccccccccc}                  
        \hline
Dataset	&	$r_{mv}$	&	MLSVM	&	MLWSVM	&	SVM	&	WSVM	&	C4.5	&	5NN	&	NB	&	LR	\\ \hline
\multirow{4}{*}{Twonorm}	&	5\%	&	0.98	&	0.98	&	0.98	&	0.98	&	0.79	&	0.97	&	0.98	&	0.98	\\
	&	10\%	&	0.98	&	0.98	&	0.98	&	0.98	&	0.78	&	0.97	&	0.97	&	0.97	\\
	&	20\%	&	0.99	&	0.99	&	0.99	&	0.99	&	0.78	&	0.97	&	0.98	&	0.98	\\
	&	40\%	&	0.97	&	0.97	&	0.97	&	0.97	&	0.78	&	0.96	&	0.97	&	0.97	\\\hline
\multirow{4}{*}{Letter}	&	5\%	&	0.99	&	1.00	&	0.99	&	1.00	&	0.99	&	1.00	&	0.97	&	0.98	\\
	&	10\%	&	0.99	&	1.00	&	0.99	&	0.99	&	0.99	&	1.00	&	0.96	&	0.99	\\
	&	20\%	&	1.00	&	1.00	&	1.00	&	1.00	&	0.99	&	1.00	&	0.96	&	0.98	\\
	&	40\%	&	1.00	&	1.00	&	1.00	&	1.00	&	0.99	&	1.00	&	0.96	&	0.98	\\\hline
\multirow{4}{*}{Ringnorm}	&	5\%	&	0.98	&	0.98	&	0.98	&	0.98	&	0.85	&	0.69	&	0.99	&	0.76	\\
	&	10\%	&	0.98	&	0.98	&	0.98	&	0.98	&	0.84	&	0.69	&	0.99	&	0.76	\\
	&	20\%	&	0.98	&	0.98	&	0.98	&	0.98	&	0.85	&	0.69	&	0.98	&	0.76	\\
	&	40\%	&	0.98	&	0.99	&	0.98	&	0.99	&	0.84	&	0.69	&	0.98	&	0.76	\\\hline
\multirow{4}{*}{Clean}	&	5\%	&	1.00	&	1.00	&	1.00	&	1.00	&	0.86	&	0.96	&	0.84	&	0.95	\\
	&	10\%	&	0.99	&	0.99	&	0.99	&	0.99	&	0.85	&	0.98	&	0.85	&	0.95	\\
	&	20\%	&	1.00	&	1.00	&	1.00	&	1.00	&	0.79	&	0.97	&	0.87	&	0.96	\\
	&	40\%	&	1.00	&	1.00	&	1.00	&	1.00	&	0.91	&	0.97	&	0.85	&	0.95	\\
\hline
\multirow{4}{*}{Forest}	&	5\%	&	0.97	&	0.98	&	0.99	&	0.99	&	0.99	&	0.99	&	0.98	&	0.98	\\
&	10\%	&	0.96	&	0.98	&	0.98	&	0.98	&	0.99	&	0.99	&	0.98	&	0.98	\\
&	20\%	&	0.98	&	0.96	&	0.97	&	0.97	&	0.99	&	0.99	&	0.98	&	0.98	\\
&	40\%	&	0.97	&	0.95	&	0.98	&	0.96	&	0.99	&	0.99	&	0.98	&	0.98	\\
\hline

\end{tabular}
\end{table*}
\normalsize

\begin{table}[h]
\footnotesize
\centering
\caption{Computational Time (sec.) for C4.5, 5NN, NB, LR, and MLSVM (excluding model selection) on public datasets. The results show that MLSVM is faster than other machine learning methods. In addition, we note that the average computational time of the REM imputation for public datasets over all missing value ratios are: Twonorm 1.22, Letter 6.89, Ringnorm 1.18, cod-rna 33.76, Clean 7.85, Advertisement 0.57, Nursery 1.41, Hypothyroid 0.16, Buzz 1705.60 sec. respectively.}
\label{tableadded4}
    \begin{tabular}{cccccc}                  
        \hline
	&	C4.5	&	5NN	&	NB	&	LR	& MLSVM \\\hline
Twonorm	&	0.79	&	0.71	&	0.46	&	0.47	&	\bf{0.35}	\\
Letter	&	0.31	&	0.69	&	0.05	&	0.06	&	\bf{0.05}	\\
Ringnorm 	&	1.29	&	1.11	&	0.91	&	0.90	&	\bf{0.58}	\\
cod-rna	&	9.88	&	9.32	&	9.04	&	9.07	&	\bf{7.87}	\\
Clean	&	4.71	&	4.01	&	3.38	&	3.78	&	\bf{3.00}	\\
Advertisement	&	10.25	&	10.05	&	16.04	&	19.91	&	\bf{9.72}	\\
Nursary 	&	0.47	&	0.77	&	0.78	&	0.89	&	\bf{0.40}	\\
Hypothyroid	&	0.18	&	0.29	&	0.22	&	0.26	&	\bf{0.15}	\\
Buzz 	&	486.04	&	580.62	&	468.17	&	479.94	&	\bf{419.92}   \\
\hline
\end{tabular}
\end{table}

\end{document}